%% file: rac4.tex
\documentclass[llncs]{article}

\newtheorem{theorem}{Theorem}

\newtheorem{proposition}{Proposition}
\newcommand{\Proof}{\NI
                    {\bf Proof.}\ }

\usepackage{url}

\def\0{{\bf 0}}
\def\1{{\bf 1}}

\include{ecmds}

\usepackage{color}
\usepackage{sgame}
\usepackage{amsmath}
\usepackage{latexsym}

\begin{document}
\date{}
\title{A comparison of the notions of optimality in 
soft constraints and graphical games}

\author{Krzysztof R. Apt$^{1,2}$, Francesca Rossi$^{3}$, and K. Brent Venable$^{3}$\\ 
\\
$^{1}$ {CWI Amsterdam, Amsterdam, The Netherlands} \\ 
$^{2}${University of Amsterdam, Amsterdam, The Netherlands} \\ 
$^{3}$
{University of Padova, Padova, Italy} \\ 
E-mails: {\tt  apt@cwi.nl}, 
{\tt \{frossi,kvenable\}@math.unipd.it}\\ \\
}

\maketitle

\begin{abstract}
 
The notion of optimality naturally arises in many areas
of applied mathematics and computer science
concerned with decision making.
Here we consider this notion in the context of two
formalisms used for different purposes and in different research areas:
graphical games and soft constraints.
We relate the notion of optimality used
in the area of soft constraint satisfaction problems (SCSPs)
to that used in graphical games,
showing that for a large class of SCSPs that includes weighted
constraints every optimal solution corresponds to a Nash equilibrium
that is also a Pareto efficient joint strategy. 

We also study alternative mappings including one that maps graphical games to
SCSPs, for which Pareto efficient joint strategies and optimal solutions coincide.

\end{abstract}

\section{Introduction}


The concept of optimality is prevalent in many areas of applied
mathematics and computer science.  It is of relevance whenever we need
to choose among several alternatives that are not equally preferable.
For example, in constraint optimization, each solution of a constraint
problem has a quality level associated with it and the aim is to
choose an optimal solution, that is, a solution with an optimal
quality level.
  
The aim of this paper is to clarify the relation between the notions
of optimality used in game theory, commonly
used to model multi-agent systems, and soft constraints.
This allows us to gain new insights into these notions which hopefully
will lead to further cross-fertilization among these two different approaches to 
model optimality.


\emph{Game theory}, notably the theory of \emph{strategic games}, captures the idea
of an interaction between agents (players). 
Each player chooses one among a set of strategies, and it
has a payoff function on the game's joint strategies that allows the player to
take action (simultaneously with the other players) with the aim of
maximizing its payoff. A commonly used concept of
optimality in strategic games is that of a Nash equilibrium. Intuitively, it is 
a joint strategy that is optimal for each player under the assumption that
only he may reconsider his action.
Another concept of optimality concerns Pareto efficient joint strategies, which 
are those in which no player can improve its payoff without 
decreasing the payoff of some other player. 
Sometimes it is useful to consider constrained Nash equilibria, that is, 
Nash equilibria that satisfy some additional requirements \cite{constNash}. 
For example, Pareto efficient Nash equilibria 
are Nash equilibria which are also Pareto efficient among the Nash equilibria.

\emph{Soft constraints}, see e.g. \cite{jacm}, 
are a quantitative formalism   
which allow us to express constraints and preferences.
While constraints state what is acceptable for a certain subset of the objects
of the problem, preferences (also called {\em soft constraints})
allow for several levels of acceptance.
An example are fuzzy constraints, see \cite{fuzzy1} and
\cite{ruttkay-fuzzy}, where acceptance levels 
are between 0 and 1, and where the quality of a solution is the
minimal level over all the constraints. An optimal
solution is the one with the highest quality.
The research in this area focuses mainly on algorithms for finding optimal solutions
and on the relationship between modelling formalisms 
(see \cite{MRS06}). 


We consider the notions of optimality in soft constraints
and in strategic games. Although apparently the only connection between 
these two formalisms is that they both model preferences, 
we show that there is in fact a strong relationship. This is surprising and interesting on its own.
Moreover, it might be exploited for a cross-fertilization among these frameworks.

In considering the relationship between
strategic games and soft constraints, 
the appropriate notion of a strategic game 
is here that of a \emph{graphical game}, see \cite{KLS01}.
This is due to the fact that soft constraints
usually involve only a small subset of the problem variables.
This is in analogy with the fact that in a graphical game a
player's payoff function depends only on a (usually small) 
number of other players.

We consider a `local' mapping that associates with each
soft constraint satisfaction problem (in short, a soft CSP, or an SCSP) a
graphical game. For strictly monotonic SCSPs
(which include, for example, weighted constraints), every optimal
solution of the SCSP is mapped to a Nash equilibrium of the game.  We also show
that this local mapping, when applied to a consistent CSP (that is, a
classical constraint satisfaction problem), maps the solutions of the CSP to the Nash
equilibria of the corresponding graphical game.  This relationship
between the optimal solutions and Nash equilibria holds in general,
and not just for a subclass, if we consider a `global' mapping from
the SCSPs to the graphical games, which is independent of the constraint
structure.

We then consider the relationship between optimal solutions of the
SCSPs and Pareto efficiency in graphical games. First we show that the above
local mapping maps every optimal
solution of a strictly monotonic SCSP to a Pareto efficient joint strategy.
We then exhibit a mapping
from the graphical games to the SCSPs for which the optimal solutions
of the SCSP coincide with the Pareto efficient joint strategies of the
game.

In \cite{gottlob} a mapping from graphical games to classical CSPs has been
defined, and it has been shown that the Nash equilibria of the games
coincide with the solutions of the CSPs.  We can use this mapping,
together with our mapping from the graphical games to the SCSPs, to
identify the Pareto efficient Nash equilibria of the given graphical
game. In fact, these equilibria correspond to the optimal solutions of
the SCSP obtained by joining the soft and hard constraints generated
by the two mappings.

The study of the relations among preference models coming from different 
fields such as AI and game theory has only recently gained attention.
In \cite{ARV05} we have considered the correspondence between optimality 
in CP-nets of \cite{BBHP.journal} and pure Nash equilibria 
in so-called parametrized strategic games, showing that there is 
a precise correspondence between these two concepts.

As mentioned above, a mapping from strategic, graphical 
and other types of games to classical CSPs 
has been considered in \cite{gottlob}, 
leading to interesting results on the complexity of deciding 
whether a game has a pure Nash equilibria or other kinds 
of desirable joint strategies.      


In \cite{tambe} a mapping from the distributed constraint optimization
problems to strategic graphical games is introduced, where the
optimization criteria is to maximize the sum of utilities.  By using
this mapping, it is shown that the optimal solutions of the given
problem are Nash equilibria of the generated game.  This result is in
line with our findings regarding strictly monotonic SCSPs, which
include the class of problems considered in \cite{tambe}.


\section{Preliminaries}
\label{back}

In this section we recall the main notions regarding 
soft constraints and strategic games.

\subsection{Soft constraints}
\label{subsec:soft}

Soft constraints, see e.g. \cite{jacm},
allow to express constraints and preferences.
While constraints state what is acceptable for a certain subset of the objects
of the problem, preferences (also called {\em soft constraints})
allow for several levels of acceptance.
A technical way to describe soft constraints is via the use of 
an algebraic structure called a c-semiring.


A \oldbfe{c-semiring} is
a tuple $\langle A,+,\times,\0,\1 \rangle$, where:
\begin{itemize}
\item $A$ is a set, called the \oldbfe{carrier} of the semiring, and $\0, \1 \in A$;
\item $+$ is commutative, associative, idempotent,
$\0$ is its unit element, and $\1$ is its absorbing element;
\item $\times$ is associative, commutative,
distributes over $+$, $\1$  is its
unit element and $\0$ is its absorbing element.
\end{itemize}

Elements $\0$ and $\1$ represent, respectively, the highest and lowest
preference.  While the operator $\times$ is used to combine
preferences, the operator + induces a partial ordering on the carrier $A$
defined by
\[
\mbox{$a \leq b$ iff $a+b = b$.}
\]

Given a c-semiring $S = \langle A,+,\times,\0,\1 \rangle$,
and a set of variables $V$, each variable $x$ with a domain $D(x)$,
a \oldbfe{soft constraint}
is a pair $\langle \text{def}, \text{con} \rangle$, where
$\text{con} \subseteq V$  and $\text{def}: \times_{y \in \text{con}} D(y) \rightarrow A$.
So a constraint specifies a set of variables (the ones in $\text{con}$),
and assigns to each tuple of values from $\times_{y \in \text{con}} D(y)$,
the Cartesian product of the variable domains, an
element of the semiring carrier $A$.

A \oldbfe{soft constraint satisfaction problem} (in short,
a \oldbfe{soft CSP} or an SCSP)
is a tuple $\langle C, V, D, S \rangle$ where $V$ is a set of variables,
with the corresponding set of domains $D$, $C$ is a set of soft constraints
over $V$ and $S$ is a c-semiring.
Given an SCSP, a \oldbfe{solution} is an instantiation
of all the variables. The \oldbfe{preference} of a solution $s$ is
the combination by means of the $\times$
operator of all the preference levels given by the constraints to the
corresponding subtuples of the solution, or more formally,
\[
\Pi_{c \in C} \text{def}_c(s \downarrow_{\text{con}_c}),
\]
where $\Pi$ is the multiplicative operator of the semiring and
$\text{def}_c(s \downarrow_{\text{con}_c})$ is the preference associated by the
constraint $c$ to the projection of the solution $s$ on the variables
in $\text{con}_c$.

A solution is called \oldbfe{optimal} if there is no other
solution with a strictly higher preference.

Three widely used instances of SCSPs are:
\begin{itemize}
\item {\bf Classical CSPs} (in short {\bf CSPs}),
based on the c-semiring
$\langle \{0,1\},\lor,\land,$ $ 0, 1\rangle$. They model
 the customary CSPs in which tuples are either allowed or not. So CSPs can be
  seen as a special case of SCSPs.
\item {\bf Fuzzy CSPs}, based on the \oldbfe{fuzzy c-semiring}
  $\langle [0,1],max,min,0,1\rangle$.
   In such problems, preferences are the values in
  $[0,1]$, combined by taking the minimum and the goal is to
  maximize the minimum preference.
\item {\bf Weighted CSPs}, based on the \oldbfe{weighted c-semiring}
$\langle\Re_{+},min,+, \infty,$ $0\rangle$.
Preferences are costs ranging over non-negative reals,
   which are aggregated using the sum. The goal is to minimize the
  total cost.
\end{itemize}




A simple example of a fuzzy CSP is the following one:
\begin{itemize}
\item three variables: $x$, $y$, and $z$, each with the domain $\{a,b\}$;
\item two constraints: $C_{xy}$ (over $x$ and $y$) and $C_{yz}$ (over
  $y$ and $z$) defined by:

$C_{xy} := \{(aa, 0.4), (ab, 0.1), (ba, 0.3), (bb, 0.5)\}$,

$C_{yz} := \{(aa, 0.4), (ab, 0.3), (ba, 0.1), (bb, 0.5)\}$.
\end{itemize}
The unique optimal solution of this problem is $bbb$ (an
abbreviation for $x=y=z=b$). Its
preference is $0.5$.

The semiring-based formalism allows one to 
model also optimization problems with several criteria.
This is done by simply considering 
SCSPs defined on c-semirings which are the Cartesian product of 
linearly ordered c-semirings. For example, the c-semiring
\[
\langle [0,1] \times [0,1], (max,max), (min,min), (\0, \0),(\1,\1)
\rangle
\]
is the Cartesian product of two fuzzy c-semirings. In a SCSP
based on such a c-semiring, preferences are pairs, e.g. (0.1,0.9),
combined using the $min$ operator on each component, e.g.
$(0.1,0.8) \times (0.3,0.6)$=$(0.1,0.6)$. The Pareto ordering induced by
using the $max$ operator on each component is a partial
ordering. In this ordering, for example,
$(0.1,0.6)<(0.2,0.8)$, while $(0.1,0.9)$ is incomparable to $(0.9,0.1)$.
More generally, if we consider the Cartesian product of $n$ 
semirings, we end up with a semiring whose elements are tuples of $n$ 
preferences, each coming from one of the given semirings. 
Two of such tuples are then ordered if each element in one of them is better 
or equal to the corresponding one in the other tuple according to the 
relevant semiring.

\subsection{Strategic games}
\label{sec:games}

Let us recall now the notion of a strategic game, see, e.g.,
\cite{Mye91}.  A strategic game for a set $N$ of $n$ players ($n > 1$) is a
sequence
\[
(S_1, \LL, S_n, p_1, \LL, p_n),
\] 
where for each $i \in [1..n]$

\begin{itemize}
\item $S_i$ is the non-empty set of \oldbfe{strategies} 
available to player $i$,

\item $p_i$ is the \oldbfe{payoff function} for the  player $i$, so
$
p_i : S_1 \times \LL \times S_n \myra A,
$
where $A$ is some fixed linearly ordered set\footnote{The use of $A$
instead of the set of real numbers precludes the construction of
mixed strategies and hence of Nash equilibria in mixed strategies, but
is sufficient for our purposes.}.
\end{itemize}

Given a sequence of non-empty
sets $S_1, \LL, S_n$ and $s \in S_1 \times \LL \times S_n$ we denote
the $i$th element of $s$ by $s_i$, abbreviate $N \setminus \{i\}$ to $-i$,
and use the following standard
notation of game theory, where $i \in [1..n]$ and $I := i_1, \LL, i_k$
is a subsequence of $1, \LL, n$:

\begin{itemize}
\item $s_{I} := (s_{i_1}, \LL, s_{i_k})$,

\item $(s'_i, s_{-i}) := (s_1, \LL, s_{i-1}, s'_i, s_{i+1}, \LL, s_n)$, where
we assume that $s'_i \in S_i$,

\item $S_{I} := S_{i_1} \times  \LL \times  S_{i_k}$.
\end{itemize}

A joint strategy $s$ is called  

\begin{itemize}
\item 
a \oldbfe{pure Nash equilibrium} (from now on, simply \oldbfe{Nash
equilibrium}) iff 
\begin{equation}
  \label{eq:nash}
p_i(s) \geq p_i(s'_i, s_{-i})  
\end{equation}
for all $i \in [1..n]$ and all $s'_i \in S_i$,

\item \oldbfe{Pareto efficient}
if for no joint strategy $s'$,
$
p_i(s') \geq p_i(s)
$
for all $i \in [1..n]$ and 
$
p_i(s') > p_i(s)
$
for some $i \in [1..n]$.
\end{itemize}

Pareto efficiency can be alternatively defined by considering
the following strict \oldbfe{Pareto ordering} $<_P$ on the $n$-tuples of reals:
\[
\mbox{$(a_1, \LL, a_n) <_P (b_1, \LL, b_n)$ iff $ \fa i \in [1..n] \ a_i \leq b_i$ 
and $\te i \in [1..n] \ a_i < b_i$.} 
\]
Then a joint strategy $s$ is Pareto efficient iff
the $n$-tuple $(p_1(s), \LL, p_n(s))$ is a maximal element in the $<_P$
ordering on such $n$-tuples of reals.

To clarify these notions consider the classical Prisoner's Dilemma
game represented by the following bimatrix representing the payoffs to
both players:
\begin{center}
\begin{game}{2}{2}
       & $C_2$    & $N_2$\\
$C_1$   &$3,3$   &$0,4$\\
$N_1$   &$4,0$   &$1,1$
\end{game}
\end{center}

Each player $i$ represents a prisoner, who
has two strategies, $C_i$ (cooperate) and $N_i$ (not cooperate).
Table entries represent payoffs for the players
(where the first component is the payoff of player 1 and the second one that of player 2).

The two prisoners gain when both cooperate (a gain of 3 each). 
However, if only one of them cooperates, 
the other one, who does not cooperate, 
will gain more (a gain of 4). If both do not cooperate, both gain very little (that is, 1 each), 
but more than the "cheated" cooperator whose cooperation is not returned (that is, 0).

Here the unique Nash equilibrium is $(N_1,
N_2)$, while the other three joint strategies $(C_1, C_2), \ (C_1,
N_2)$ and $(N_1, C_2)$ are Pareto efficient.

\subsection{Graphical games}
\label{graph-g}

A related modification of the concept of strategic games, called
\oldbfe{graphical games}, was proposed in \cite{KLS01}. These games
stress the locality in taking decision.  
In a graphical game the
payoff of each player depends only on the strategies of its neighbours
in a given in advance graph structure over the set of players.

More formally, a \oldbfe{graphical game} for $n$ players with the corresponding
strategy sets $S_1, \LL, S_n$ with the payoffs being elements of a linearly ordered
set $A$, is defined by assuming a neighbour
function \emph{neigh} that given a player $i$ yields its set of
neighbours $\emph{neigh}(i)$. The payoff for player $i$ is then a
function $p_i$ from $\Pi_{j \in \emph{neigh}(i) \cup \{i\}} S_j$ to $A$.  
We denote such a graphical game by
\[
(S_1, \dots, S_n, \emph{neigh}, p_1, \dots, p_n, A).
\]

By using the canonical extensions of these payoff functions to the
Cartesian product of all strategy sets one can then extend the
previously introduced concepts to the graphical games.  Further, when
all pairs of players are neighbours, a graphical game reduces to a
strategic game.

\section{Optimality in SCSPs and Nash equilibria in graphical games}
\label{sec:soft-to}

In this section we relate the notion of optimality 
in soft constraints and the concept of Nash equilibria in graphical games. 
We shall see that, while CSPs are sufficient to 
obtain the Nash equilibria of any given graphical game, 
the opposite direction does not hold. However, graphical games 
can model, via their Nash equilibria, a superset  
of the set of the optimal solutions of any given SCSP.


The first statement is based on a result in \cite{gottlob}, where,
given a graphical game, it is shown how to build a corresponding CSP
such that the Nash equilibria of the game and the solutions of the CSP coincide.
Thus, the full expressive power of SCSPs is not needed to 
model the Nash equilibria of a game.
We will now focus on the opposite direction: from SCSPs 
to graphical games. Unfortunately, the inverse of the mapping defined in 
\cite{gottlob} cannot be used for this purpose since it only returns 
CSPs of a specific kind. 

\subsection{From SCSPs to graphical games: a local mapping}
\label{local}

We now define a mapping from soft CSPs to
a specific kind of graphical games.
We identify the players with the variables. Thus,  
since soft constraints link variables, the
resulting game players are naturally connected.  To capture this
aspect, we use graphical games. 
We allow here payoffs to be elements of an arbitrary
linearly ordered set.


Let us consider a first possible mapping from SCSPs to graphical games.
In what follows we focus on SCSPs based on c-semirings with the carrier
linearly ordered by $\leq$ (e.g.~fuzzy or weighted) and on the concepts of
optimal solutions in SCSPs and Nash equilibria in graphical games.

Given a SCSP $P := \langle C, V, D, S \rangle$ we define the corresponding 
graphical game for $n=|V|$ players as follows:

\begin{itemize}
\item the players: one for each variable;
\item the strategies of player $i$: all values in the domain of the
  corresponding variable $x_i$;

\item the neighbourhood relation: $j \in \emph{neigh}(i)$ iff the variables
$x_i$ and $x_j$ appear together in some constraint from $C$;

\item the payoff function of player $i$:
   
  Let $C_i \subseteq C$ be the set of constraints involving $x_i$ and
  let $X$ be the set of variables that appear together with $x_i$ in
  some constraint in $C_i$ (i.e., $X = \C{x_j \mid j \in \emph{neigh}(i)}$).
  Then given an assignment $s$ 
to all variables in $X \cup \C{x_i}$ the payoff
  of player $i$ w.r.t.~$s$ is defined by:
  
\[
p_i(s) :=\Pi_{c \in C_i} \text{def}_c(s \downarrow_{\text{con}_c}).
\]
\end{itemize}

We denote the resulting graphical game by $L(P)$ to emphasize 
the fact that the payoffs are
obtained using \emph{local} information about each variable, by looking
only at the constraints in which it is involved.   

One could think of a different mapping where players correspond 
to constraints. However, such a mapping can be obtained by applying the 
local mapping $L$ to the hidden variable encoding \cite{hidden} of the SCSP in input. 

We now analyze the relation between the optimal solutions of a SCSP
$P$ and the Nash equilibria of the 
derived game $L(P)$.  

\subsubsection{General case}

In general, these two concepts are unrelated.
Indeed, consider the fuzzy CSP defined 
at the end of Section \ref{subsec:soft}.
The corresponding game has:

\begin{itemize}
\item three players, $x$, $y$, and $z$;
\item each player has two strategies, $a$ and $b$;
\item the neighbourhood relation is defined by:
\[
\emph{neigh}(x):=\{y \}, \ \emph{neigh}(y):=\{x,z \}, \ \emph{neigh}(z):=\{ y\};
\]
\item the payoffs of the players are defined as follows:
\begin{itemize}
\item for player $x$: 

$p_x(aa*):=0.4$, $p_x(ab*):=0.1$, $p_x(ba*):=0.3$, $p_x(bb*):=0.5$;

\item for player $y$: 

$p_y(aaa):=0.4$, $p_y(aab):=0.3$, $p_y(abb):=0.1$, $p_y(bbb):=0.5$, 

$p_y(bba):=0.5$, $p_y(baa):=0.3$, $p_y(bab):=0.3$, $p_y(aba):=0.1$;
     
\item for player $z$: 

$p_z(*aa):=0.4$, $p_z(*ab):=0.3$, $p_z(*ba):=0.1$, $p_z(*bb):=0.5$;
\end{itemize}
\end{itemize}
where $*$ stands for either $a$ or $b$ and where to facilitate the
analysis we use the canonical extensions of the payoff functions
$p_x$ and $p_z$ to the functions on $\C{a,b}^3$.

This game has two Nash equilibria: $aaa$ and $bbb$. However, only
$bbb$ is an optimal solution of the fuzzy SCSP.

One could thus think that in general
the set of Nash equilibria is a superset of the
set of optimal solutions of the corresponding 
SCSP. However, this is not the case. Indeed, 
consider a fuzzy CSP with as before three variables, $x, y$ and $z$, 
each with the domain $\C{a,b}$, but now with the constraints:
\II

$C_{xy} := \{(aa, 0.9), (ab, 0.6), (ba, 0.6), (bb, 0.9)\}$,

$C_{yz} := \{(aa, 0.1), (ab, 0.2), (ba, 0.1), (bb, 0.2)\}$.
\II

Then $aab, \ abb, \ bab$ and $bbb$ are all optimal solutions but only
$aab$ and $bbb$ are Nash equilibria of the corresponding graphical game.

\subsubsection{SCSPs with strictly monotonic combination} 
\label{subsub:mon}

Next, we consider the case when the multiplicative operator $\times$ is
strictly monotonic.  Recall that given a c-semiring $\langle A,+,
\times, \0, \1 \rangle$, the operator $\times$ is \oldbfe{strictly
  monotonic} if for any $a,b,c \in A$ such that $a<b$ we have $c
\times a < c \times b$.  (The symmetric condition is taken care of by
the commutativity of $\times$.)  

Note for example that in the case of
classical CSPs $\times$ is not strictly monotonic, as $a < b$ implies
that $a = 0$ and $b = 1$ but $c \land a < c \land b$ does not
hold then for $c = 0$.  Also in fuzzy CSPs $\times$ is not strictly
monotonic, as $a < b$ does not imply that $min(a,c) < min(b,c)$ for
all $c$.  In contrast, in weighted CSP $\times$ is strictly monotonic,
as $a < b$ in the carrier means that $b < a$ as reals, so for any $c$
we have $c + b < c + a$, i.e., $c \times a < c \times b$ in the carrier.

So consider now a c-semiring with a linearly ordered carrier and a 
strictly monotonic multiplicative operator. As in the previous case,
given an SCSP $P$, it is possible that a Nash equilibrium of $L(P)$ is
not an optimal solution of $P$. Consider for example a
weighted SCSP $P$ with
\begin{itemize}
\item two variables, $x$ and $y$, each with the
domain $D=\{a,b\}$;
\item one constraint $C_{xy} := \{(aa, 3), (ab, 10), (ba, 10), (bb, 1)\}$.
\end{itemize}
The corresponding game $L(P)$ has:

\begin{itemize}
\item two players, $x$ and $y$, who are neighbours of each other;

\item each player has two strategies, $a$ and $b$;

\item the payoffs defined by: 

$p_x(aa):=p_y(aa):=7$, $p_x(ab):=p_y(ab):=0$, 

$p_x(ba):=p_y(ba):=0$, $p_x(bb):=p_y(bb):=9$.
\end{itemize}

Notice that, in a weighted CSP we have $a \leq b$ in the carrier iff
$b \leq a$ as reals, so when passing from the SCSP to the
corresponding game, we have complemented the costs w.r.t.~10, when
making them payoffs. In general, given a weighted CSP, we can define
the payoffs (which must be maximized) from the costs (which must be
minimized) by complementing the costs w.r.t.~the greatest cost used in
any constraint of the problem.

Here $L(P)$ has two Nash equilibria, $aa$ and $bb$, but only $bb$ is
an optimal solution.  Thus, as in the fuzzy case, we have that there
can be a Nash equilibrium of $L(P)$ that is not an optimal solution of
$P$.  However, in contrast to the fuzzy case, when the multiplicative
operator of the SCSP is strictly monotonic, the set of Nash equilibria
of $L(P)$ is a superset of the set of optimal solutions of $P$.

\begin{theorem}
Consider a SCSP $P$ defined on a c-semiring $\langle A,+,\times,\0,\1
\rangle$, where $A$ is linearly ordered and $\times$ is 
strictly monotonic, and the corresponding game $L(P)$. 
Then every optimal solution of
$P$ is a Nash equilibrium of $L(P)$.  
\end{theorem}

\Proof We prove that if a joint strategy $s$ 
is not a Nash equilibrium of game $L(P)$, then it is not an optimal 
solution of SCSP $P$. 

Let $a$ be the strategy of player $x$ in $s$, and let
$s_{\emph{neigh}(x)}$ and $s_{Y}$ 
be, respectively, the joint strategy of the neighbours of $x$, and of
all other players, in $s$. That is, $V = \{x\} 
\cup \emph{neigh}(x) \cup Y$ and we write $s$ as
$(a, s_{\emph{neigh}(x)}, s_Y)$.

By assumption there is a strategy $b$ for $x$ such that the payoff 
$p_{x}(s')$ for the joint strategy 
$s' := (b, s_{\emph{neigh}(x)}, s_Y)$ is higher than 
$p_x(s)$. (We use here the canonical extension of $p_x$ to the Cartesian product of
all the strategy sets).

So by the definition of the mapping $L$
\[
\Pi_{c \in C_x} \text{def}_c(s \downarrow_{\text{con}_c})< \Pi_{c \in C_x}
\text{def}_c(s' \downarrow_{\text{con}_c}),
\]
where $C_x$ is the set of all the constraints involving $x$ in SCSP $P$. 
But the preference of $s$ and $s'$ is the same on all the
constraints not involving $x$ and $\times$ is strictly monotonic,
so we conclude that
\[
\Pi_{c \in C} \text{def}_c(s \downarrow_{\text{con}_c})< \Pi_{c \in C}
\text{def}_c(s' \downarrow_{\text{con}_c}).
\]
This means that $s$ is not an optimal solution of $P$.
\HB

\subsubsection{Classical CSPs}

The above result does not hold for classical CSPs.
Indeed, consider a CSP with:
\begin{itemize}
\item three variables: $x$, $y$, and $z$, each with the domain $\{a,b\}$;
\item two constraints: $C_{xy}$ (over $x$ and $y$) and $C_{yz}$ (over
  $y$ and $z$) defined by:

$C_{xy} := \{(aa, 1), (ab, 0), (ba, 0), (bb, 0)\}$,

$C_{yz} := \{(aa, 0), (ab, 0), (ba, 1), (bb, 0)\}$.
\end{itemize}

This CSP has no solutions in the classical sense, i.e., each optimal solution, in particular $baa$, 
has preference 0. However,
$baa$ is not a Nash equilibrium of the resulting graphical game, since
the payoff of player $x$ increases when he switches to the strategy $a$.

On the other hand, if we
restrict the domain of $L$ to consistent CSPs, that is, CSPs with at least one
solution with value 1, then the discussed inclusion does hold.

\begin{proposition}
Consider a consistent CSP $P$ and the corresponding game $L(P)$. 
Then every solution of $P$ is a Nash equilibrium of $L(P)$.  
\end{proposition}
\Proof 
Consider a solution $s$ of $P$. In the resulting game $L(P)$ the payoff
to each player is maximal, namely 1. So the 
joint strategy $s$ is a Nash equilibrium in game $L(P)$.
\HB 
\VV

The reverse inclusion does not need to hold.
Indeed, consider the following CSP:
\begin{itemize}
\item three variables: $x$, $y$, and $z$, each with the domain $\{a,b\}$;
\item two constraints: $C_{xy}$ and $C_{yz}$  defined by:

$C_{xy} := \{(aa, 1), (ab, 0), (ba, 0), (bb, 0)\}$,

$C_{yz} := \{(aa, 1), (ab, 0), (ba, 0), (bb, 0)\}$.

\end{itemize}
Then $aaa$ is a solution, so the CSP is consistent. But
$bbb$ is not an optimal solution, while it
is a Nash equilibrium of the resulting game.

So for consistent CSPs our mapping $L$ yields games in which the set
of Nash equilibria is a, possibly strict, superset of the set of
solutions of the CSP.

However, there are ways to relate CSPs and games so that the 
solutions and the Nash equilibria coincide. This is what is done in 
\cite{gottlob}, where the mapping is from the strategic games to CSPs. 
Notice that our mapping goes in the opposite direction and 
it is not the reverse of the one in \cite{gottlob}. In fact, the mapping
in \cite{gottlob} is not reversible.

\subsection{From SCSPs to graphical games: a global mapping}
\label{global}

Other mappings from SCSPs to games can be defined. 
While our mapping $L$ is in some sense `local', since it considers
the neighbourhood of each variable, we can also define 
an alternative `global' mapping that considers all constraints. 
More precisely, 
given a SCSP $P=\langle C, V, D, S \rangle$, with a linearly ordered carrier $A$ of $S$,
we define the corresponding 
game on $n=|V|$ players, 
$GL(P)=(S_1, \dots, S_n, p_1, \dots, p_n, A)$
by using the following payoff function $p_i$ for player $i$:

\begin{itemize}

\item 
given an assignment $s$ to \emph{all} variables in $V$
\[
p_i(s) := \Pi_{c \in C} \text{def}_c(s \downarrow_{\text{con}_c}).
\]
\end{itemize}

Notice that in the resulting game the payoff functions of all 
players are the same.

\begin{theorem}
\label{10}
Consider an SCSP $P$ over a linearly ordered carrier,
and the corresponding game $GL(P)$. 
Then every optimal solution of $P$ is a Nash equilibrium of $GL(P)$.  
\end{theorem}

\Proof
An optimal solution of $P$, say $s$, 
is a joint strategy for which all players have the same, highest, payoff.
So no other joint strategy exists for which some 
player is better off and consequently $s$ is a Nash equilibrium.
\HB 
\VV

The opposite inclusion does not need to hold.
Indeed, consider again the weighted SCSP of Subsection \ref{subsub:mon}
with 
\begin{itemize}
\item two variables, $x$ and $y$, each with the domain $D=\{a,b\}$;

\item one constraint, $C_{xy} := \{(aa, 3), (ab, 10), (ba, 10), (bb, 1)\}$.
\end{itemize}
Since there is one constraint, the mappings $L$ and $GL$ coincide.
Thus we have that $aa$ is a Nash equilibrium of $GL(P)$ but is not 
an optimal solution of $P$.

While the mapping defined in this section has the advantage of 
providing a precise subset relationship between 
optimal solutions and Nash equilibria, as Theorem 
\ref{10} states, it has an obvious disadvantage 
from the computational point of view, since it requires
to consider all the complete assignments of the SCSP.

\subsection{Summary of results}

Summarizing, in this section we have analyzed the relationship between 
the optimal solutions of SCSPs and the Nash equilibria of graphical games. 
In \cite{gottlob} CSPs have been shown to be sufficient to model Nash equilibria of graphical games. 
Here we have considered the question whether the Nash equilibria of graphical games can model the 
optimal solutions of SCSPs. We have provided two mappings from SCSPs to graphical games,
showing that (with some conditions for the local mapping) 
the set of Nash equilibria of the obtained game contains the optimal solutions of the given SCSP.

Nash equilibria can be seen as the optimal elements in very specific orderings,
where dominance is based on exactly one change in the joint strategy,
while SCSPs can model any ordering. So we conjecture that it is not 
possible to find a mapping from SCSPs to the graphical games for which the optimals coincide with Nash equilibria.
Such a conjecture is also supported by the fact that strict Nash equilibria 
can be shown to coincide with the optimals of a CP-net, see \cite{ARV05}, 
and the CP-nets can model strictly less orderings than the SCSPs, see \cite{prest}. 
 
\section{Optimality in SCSPs and Pareto efficient joint strategies in graphical games}
\label{sec:to-soft}








Next, we relate the notion of optimality in SCSPs to the Pareto efficient joint strategies of 
graphical games. 

\subsection{From SCSPs to graphical games}

Consider again the local and the global mappings from SCSPs 
to graphical games 
defined in Sections \ref{local} and \ref{global}.
We will now prove that the local mapping yields a 
game whose set of Pareto efficient joint strategies
contains the set of optimal solutions of a given SCSP. 
On the other hand, the global mapping 
gives a one-to-one correspondence between the two sets.

\begin{theorem}
Consider an SCSP $P$ defined on a c-semiring $\langle A,+,\times,\0,\1
\rangle$, where $A$ is linearly ordered and $\times$ is strictly monotonic, 
and the corresponding game $L(P)$.
Then every optimal solution of $P$ is a Pareto efficient joint strategy
of $L(P)$.
\end{theorem}

\Proof
Let us consider a joint strategy $s$ of L(P) which is not Pareto efficient.
We will show that $s$ does not correspond to an optimal solution of $P$.
Since $s$ is not Pareto efficient, there is a joint strategy 
$s'$ such that $p_i(s) \leq p_i(s')$ for all $i \in [1..n]$ and 
$p_i(s) < p_i(s')$ for some $i \in [1..n]$.
Let us denote with $I = \{i \in [1..n]$ such that $p_i(s) < p_i(s')\}$.
By the definition of the mapping $L$, we have:
\[
\Pi_{c \in C_i} \text{def}_c(s \downarrow_{\text{con}_c})< \Pi_{c \in C_i}
\text{def}_c(s' \downarrow_{\text{con}_c}),
\]
for all $i \in I$ and where $C_i$ is the set of all 
the constraints involving the variable corresponding to player $i$ in SCSP $P$. 
Since the preference of $s$ and $s'$ is the same on all the
constraints not involving any $i \in I$, and since $\times$ is strictly monotonic,
we have:
\[
\Pi_{c \in C} \text{def}_c(s \downarrow_{\text{con}_c})< \Pi_{c \in C}
\text{def}_c(s' \downarrow_{\text{con}_c}).
\]
This means that $s$ is not an optimal solution of $P$.
\HB \VV

To see that there may be Pareto efficient joint strategies that do not correspond to the 
optimal solutions, consider a
weighted SCSP $P$ with
\begin{itemize}
\item two variables, $x$ and $y$, each with 
domain $D=\{a,b\}$;
\item constraint $C_{x} := \{(a, 2), (b, 1)\}$;
\item constraint $C_{y} := \{(a, 4), (b, 7)\}$;
\item constraint $C_{xy} :=\{(aa,0), (ab, 10), (ba, 10), (bb,0) \}$.
\end{itemize}
The corresponding game $L(P)$ has:

\begin{itemize}
\item two players, $x$ and $y$, who are neighbours of each other;

\item each player has two strategies: $a$ and $b$;

\item the payoffs defined by: 
$p_x(aa):=8$, 
$p_y(aa):=6$, 
$p_x(ab):=p_y(ab):=0$, 
$p_x(ba):=p_y(ba):=0$, 
$p_x(bb):= 9$,
$p_y(bb):=3$.
\end{itemize}

As in Section \ref{local} 
when passing from an SCSP to the
corresponding game, we have complemented the costs w.r.t.~10, when
turning them to payoffs. 
$L(P)$ has two Pareto efficient joint strategies: $aa$ and $bb$. 
(They are also both Nash equilibria.) However, only 
$aa$ is optimal in $P$.

If the combination operator is idempotent, there is no relation 
between the optimal solutions of $P$ and the Pareto efficient joint strategies 
of $L(P)$.  
However, if we use the global mapping defined in Section \ref{global},  the
optimal solutions do correspond to Pareto efficient joint strategies,
regardless of the type of the combination operator. 

\begin{theorem}
Consider an SCSP $P$ defined on a c-semiring $\langle A,+,\times,\0,\1
\rangle$, where $A$ is linearly ordered, 
and the corresponding game $GL(P)$.
Then every optimal solution of $P$ is a Pareto efficient joint strategy
of $GL(P)$, and viceversa.
\end{theorem}

\Proof
Any optimal solution corresponds to a joint strategy where all players have  
the same payoff, which is the solution's preference. 
Thus, such a joint strategy cannot be Pareto dominated by any other strategy.
Conversely, a solution corresponding to a joint strategy with the highest 
payoff is optimal. 
\HB

\subsection{From graphical games to SCSPs}

Next, we define a mapping from graphical games to SCSPs
that relates Pareto efficient joint strategies
in games to optimal solutions in SCSPs.
In order to define such a mapping,
we limit ourselves to SCSPs defined on c-semirings which 
are the Cartesian product of linearly ordered c-semirings (see Section \ref{subsec:soft}).
More precisely, given a graphical game $G=(S_1, \dots, S_n, \emph{neigh}, p_1, \dots, p_n, A)$
we define the corresponding SCSP $L'(G)= \langle C, V, D, S \rangle$, 
as follows:
\begin{itemize}

\item each variable $x_i$ corresponds to a player $i$;

\item the domain $D(x_i)$ of the variable $x_i$ consists of
  the set of strategies of player $i$, i.e., $D(x_i) : = S_i$;

\item the c-semiring is 

$\langle A_1 \times \cdots \times A_n, (+_1, \dots, +_n),
(\times_1, \dots, \times_n),(\0_1, \dots, \0_n),(\1_1, \dots, \1_n) \rangle$,

\NI
the Cartesian product of $n$ \emph{arbitrary} linearly ordered semirings;

\item soft constraints: for each variable $x_i$, one constraint $\langle
  \text{def},\text{con} \rangle$ such that:
\begin{itemize}
\item $\text{con} = \emph{neigh}(x_i) \cup \{x_i \}$;
\item $\text{def}: \Pi_{y \in \text{con}} D(y) \rightarrow  A_1 \times \cdots
  \times A_n$ such that for any $s \in \Pi_{y \in \text{con}} D(y)$, 
  $\text{def}(s) :=(d_1, \dots, d_n)$ with $d_j=\1_j$ for every $j \neq i$ and 
  $d_i=f(p_i(s))$, where 
$f: A \rightarrow A_i$ is an order preserving
mapping from payoffs to preferences
(i.e., if $r>r'$ then $f(r)>f(r')$ in the c-semiring's ordering). 
\end{itemize}
\end{itemize}  

To illustrate it consider again the previously used Prisoner's Dilemma game:

\begin{center}
\begin{game}{2}{2}
       & $C_2$    & $N_2$\\
$C_1$   &$3,3$   &$0,4$\\
$N_1$   &$4,0$   &$1,1$
\end{game}
\end{center}

Recall that in this game the only Nash equilibrium is 
$(N_1,N_2)$, while the other three joint strategies are 
Pareto efficient.

We shall now construct a corresponding SCSP based on 
the Cartesian product of two weighted semirings.
This SCSP according to the mapping $L'$ has:\footnote{Recall that in the weighted semiring
\textbf{1} equals 0.}
\begin{itemize}
\item two variables: $x_1$ and $x_2$, each with the domain
$\{c,n\}$;
\item two constraints, both on $x_1$ and $x_2$:
\begin{itemize}
\item constraint $c_1$ with 
$\text{def}(cc) := \langle 7, 0 \rangle$, 
$\text{def}(cn) := \langle 10, 0 \rangle$, 
$\text{def}(nc) := \langle 6, 0 \rangle$, 
$\text{def}(nn) := \langle 9, 0 \rangle$;
\item constraint $c_2$ with 
$\text{def}(cc) := \langle 0, 7 \rangle$, 
$\text{def}(cn) := \langle 0, 6 \rangle$, 
$\text{def}(nc) := \langle 0, 10 \rangle$, 
$\text{def}(nn) := \langle 0, 9 \rangle$;
\end{itemize}
\end{itemize}

The optimal solutions of this SCSPs are: $cc$, with preference
$\langle 7,7 \rangle$, $nc$, with preference $\langle 10,6 \rangle$,
$cn$, with preference $\langle 6,10 \rangle$.  The remaining solution,
$nn$, has a lower preference in the Pareto ordering. Indeed, its
preference $\langle 9,9 \rangle$ is dominated by $\langle 7,7 \rangle$, 
the preference of $cc$ (since
preferences are here costs and have to be minimized).  Thus the
optimal solutions coincide here with the Pareto efficient joint strategies
of the given game. This is true in general.

\begin{theorem}
\label{t1}
Consider a graphical game $G$ and a corresponding SCSP $L'(G)$.
Then the optimal solutions of $L'(G)$ coincide with 
the Pareto efficient joint strategies of $G$.
\end{theorem}

\Proof
In the definition of the mapping $L'$ we stipulated that the mapping
$f$ maintains the ordering from the payoffs to preferences.
As a result
each joint strategy $s$ corresponds to the $n$-tuple 
of preferences $(f(p_1(s)), \dots, f(p_n(s)))$
and the Pareto orderings on the $n$-tuples  $(p_1(s), \dots, p_n(s))$
and $(f(p_1(s)), \dots,$ $f(p_n(s)))$ coincide.
Consequently a sequence $s$
is an optimal solution of the SCSP $L'(G)$ iff
$(f(p_1(s)), \dots, f(p_n(s)))$ is a maximal element 
of the corresponding Pareto ordering. 
\HB \VV

We notice that $L'$ is injective and, thus, can be reversed on its image.
When such a reverse mapping is applied to these specific SCSPs,
payoffs correspond to projecting of the players' valuations to a subcomponent.


\subsubsection{Pareto efficient Nash equilibria}

As mentioned earlier, in \cite{gottlob} a mapping is defined
from the graphical
games to CSPs such that Nash equilibria coincide with the solutions of CSP.  
Instead, our mapping is from the graphical games to
SCSPs, and is such that Pareto efficient joint strategies
and the optimal solutions coincide.

Since CSPs can be seen as a special instance of SCSPs, where only
\textbf{1}, \textbf{0}, the top and bottom elements of the semiring,
are used, it is possible to add to any SCSP a set of hard constraints.
Therefore we can merge the results of the two mappings into a single 
SCSP, which contains the soft constraints generated by $L'$ 
and also the hard constraints generated by the mapping in \cite{gottlob},
Below we denote these hard constraints by $H(G)$.
We recall that 
each constraint in $H(G)$ corresponds to a player, 
has the variables corresponding 
to the player and it neighbours and allows only tuples corresponding to the
strategies in which the player has no so-called regrets.
If we do this, then the optimal solutions of the new SCSP
with preference higher than \textbf{0}
are the Pareto efficient Nash equilibria of the given game, that is, 
those Nash equilibria which dominate or are 
incomparable with all other Nash equilibria according  
to the Pareto ordering. Formally, we have the following result.
  
\begin{theorem}
Consider a graphical game $G$ and the SCSP $L'(G) \cup H(G)$.
If the optimal solutions of $L'(G) \cup H(G)$
have global preference greater than \textbf{0},
they correspond to the 
Pareto efficient Nash equilibria of $G$.
\end{theorem}

\Proof 
Given any solution $s$, let $p$ be its preference in $L'(G)$
and $p'$ in $L'(G) \cup H(G)$. By the construction of the constraints
$H(G)$ we have that $p'$ equals $p$ if $s$ is a Nash equilibrium and
$p'$ equals \textbf{0} otherwise.  The remainder of the argument is as
in the proof of Theorem \ref{t1}.  \HB \VV

For example, in the Prisoner's Dilemma game,
the mapping in \cite{gottlob} would generate just one constraint
on $x_1$ and $x_2$ with $nn$ as the only allowed tuple.
In our setting, when using as the linearly ordered c-semirings
the weighted semirings,
this would become a soft constraint
with 
\[
\text{def}(cc) := \text{def}(cn) := \text{def}(nc) = \langle \infty, \infty \rangle, \
\text{def}(nn) := \langle 0, 0 \rangle.
\]
With this new constraint, all solutions have the preference 
$\langle \infty, \infty \rangle$, except for $nn$ which has the
preference $\langle 9,9 \rangle$ and thus is optimal.  
This solution corresponds to the joint strategy $(N_1,N_2)$ with the payoff $(1,1)$ 
(and thus preference $(9,9)$).
This is the only Nash equilibrium and thus the only Pareto efficient Nash equilibrium. 

 
This method allows us to identify among Nash equilibria the `optimal' ones.
One may also be interested in knowing whether there exist
Nash equilibria which are also Pareto efficient joint strategies.
For example, in the Prisoners' Dilemma example, there are no such Nash equilibria.
To find any such joint strategies we can use the two mappings separately, 
to obtain, given a game $G$, both an SCSP $L'(G)$ and a CSP $H(G)$
(using the mapping in \cite{gottlob}). Then we should take the intersection 
of the set of optimal solutions of $L'(G)$ and the set of solutions of 
$H(G)$. 

\subsection{Summary of results}

We have considered the relationship between optimal solutions of SCSPs
and Pareto efficient joint strategies in graphical games.  The local mapping of Section
\ref{local} turns out to map optimal solutions of a given SCSP to
Pareto efficient joint strategies, while the global mapping of Section \ref{global}
yields a one-to-one correspondence.  For the reverse direction it is
possible to define a mapping such that these two notions of optimality
coincide. However, none of these mappings are onto.
  
\section{Conclusions}
\label{conc}

In this paper we related two formalisms that are commonly used to
reason about optimal outcomes: graphical games and soft
constraints. While for soft constraints there is only one notion of optimality, for 
graphical games there are at least two. In this paper we have considered 
Nash equilibria and Pareto efficient joint strategies.

We have defined a natural mapping from SCSPs that combine preferences
using a strictly monotonic operator to a class of graphical games such
that the optimal solutions of the SCSP are included in the Nash
equilibria of the game and in the set of Pareto efficient joint strategies.  In general
the inclusions cannot be reversed.  We have also exhibited a mapping
from the graphical games to a class of SCSPs such that the Pareto efficient joint strategies
of the game coincide with the optimal solutions of the SCSP.



These results can be used in many ways. One obvious way
is to try to exploit computational 
and algorithmic results existing for one of these areas in another.
This has been pursued already in \cite{gottlob} for games by using
hard constraints. As a consequence of our results this can also be
done for strategic games by using soft constraints.  For example,
finding a Pareto efficient joint strategy involves mapping a game into
an SCSP and then solving it. A similar approach can also be applied to
Pareto efficient Nash equilibria, which can be found by solving a
suitable SCSP.

\bibliographystyle{plain}



\end{document}

%% file: ecmds.tex
\newcommand{\oldbfe}[1]{\begin{bfseries}\emph{#1}\end{bfseries}}

\newcommand{\myra}{\mbox{$\:\rightarrow\:$}}

\newcommand{\fa}{\mbox{$\forall$}}
\newcommand{\te}{\mbox{$\exists$}}

\newcommand{\LL}{\mbox{$\ldots$}}

\newcommand{\C}[1]{\mbox{$\{{#1}\}$}}           

\newcommand{\NI}{\noindent}
\newcommand{\HB}{\hfill{$\Box$}}
\newcommand{\VV}{\vspace{5 mm}}

\newcommand{\II}{\vspace{2 mm}}

\newcommand{\szkew}[1]{\relax \setbox0=\hbox{\kern -24pt $\displaystyle#1$\kern 0pt }%
\box0}
{\catcode`\@=11 \global\let\ifjusthvtest@=\iffalse}

\newcounter{oldmycaption}













%% file: rac4.bbl
\begin{thebibliography}{10}


\bibitem{ARV05}
K.R. Apt, F.~Rossi, and K.~B. Venable.
\newblock CP-nets and Nash equilibria.
\newblock In {\em Proc. of the Third International Conference on Computational
  Intelligence, Robotics and Autonomous Systems (CIRAS '05)}, pages 1--6.
\newblock Available from \url{http://arxiv.org/abs/cs/0509071}.

\bibitem{jacm}
S.~Bistarelli, U.~Montanari, and F.~Rossi.
\newblock Semiring-based constraint solving and optimization.
\newblock {\em Journal of the ACM}, 44(2):201--236, mar 1997.

\bibitem{BBHP.journal}
C.~Boutilier, R.~I. Brafman, C.~Domshlak, H.~H. Hoos, and D.~Poole.
\newblock {CP}-nets: A tool for representing and reasoning with conditional
  ceteris paribus preference statements.
\newblock {\em J. Artif. Intell. Res. (JAIR)}, 21:135--191, 2004.



\bibitem{fuzzy1}
H.~Fargier D.~Dubois and H.~Prade.
\newblock The calculus of fuzzy restrictions as a basis for flexible constraint
  satisfaction.
\newblock In {\em IEEE International Conference on Fuzzy Systems}, 1993.


\bibitem{gottlob}
G.~Greco G.~Gottlob and F.~Scarcello.
\newblock Pure {Nash} equilibria: hard and easy games.
\newblock {\em J. of Artificial Intelligence Research}, 24:357--406, 2005.



\bibitem{constNash}
G.~Greco and F.~Scarcello.
\newblock Constrained pure Nash equilibria in graphical games,
\newblock Proceedings of the 16th Eureopean Conference on Artificial
               Intelligence (ECAI'2004), pages 181--185, IOS Press, 2004.  

\bibitem{KLS01}
M.~Kearns, M.~Littman, and S.~Singh.
\newblock Graphical models for game theory.
\newblock In {\em Proceedings of the 17th Conference in Uncertainty in
  Artificial Intelligence (UAI '01)}, pages 253--260. Morgan Kaufmann, 2001.

\bibitem{Mye91}
R.~B. Myerson.
\newblock {\em Game Theory: Analysis of Conflict}.
\newblock Harvard Univ Press, Cambridge, Massachusetts, 1991.


\bibitem{MRS06}
F.~Rossi P.~Meseguer and T.~Schiex.
\newblock Soft constraints.
\newblock In T.~Walsh F.~Rossi, P. Van~Beek, editor, {\em Handbook of
  Constraint programming}, pages 281--328. Elsevier, 2006.


\bibitem{prest} 
C. Domshlak, S. Prestwich, F. Rossi, K. B. Venable, T. Walsh.
\newblock Hard and soft constraints 
for reasoning about qualitative conditional preferences.
\newblock In {\em Journal of Heuristics}, Special issue on preferences, 12:
263-285, Springer, 2006.


\bibitem{ruttkay-fuzzy}
Z.~Ruttkay.
\newblock Fuzzy constraint satisfaction.
\newblock In {\em Proceedings 1st {IEEE} Conference on Evolutionary Computing},
  pages 542--547, Orlando, 1994.


\bibitem{tambe}
R. T.~Maheswaran, J. P.~Pearce, and M.~Tambe. 
\newblock Distributed algorithms for DCOP: a graphical-game-based approach. 
\newblock In Proceedings of the ISCA 17th International Conference on
Parallel and Distributed Computing Systems (ISCA PDCS 2004), pages 432--439, ISCA, 2004.

\bibitem{hidden}
N. Mamoulis and K. Stergiou.
\newblock Solving non-binary CSPs using the hidden variable encoding.
\newblock In Lecture Notes in Computer Science volume 2239, Springer, 2001.




\end{thebibliography}
